\documentclass[10pt,twocolumn,letterpaper]{article}

\usepackage[pagenumbers]{cvpr} 

\usepackage{graphicx}
\usepackage{amsmath}
\usepackage{amssymb}
\usepackage{booktabs}

\usepackage{bm}
\usepackage{url}
\usepackage{pifont}
\usepackage{bbding}
\usepackage{comment}
\usepackage{caption}
\usepackage{multirow}
\usepackage{makecell}
\usepackage{amsfonts}
\usepackage[misc]{ifsym}
\usepackage{pifont}
\usepackage{xcolor,colortbl}
\newcommand{\cmark}{\ding{51}}%

\def\ie{{\em i.e. }}
\def\eg{{\em e.g. }}
\def\etal{{\em et al. }}

\def\conv{AeConv}

\usepackage[pagebackref,breaklinks,colorlinks]{hyperref}

\usepackage[capitalize]{cleveref}
\crefname{section}{Sec.}{Secs.}
\Crefname{section}{Section}{Sections}
\Crefname{table}{Table}{Tables}
\crefname{table}{Tab.}{Tabs.}

\def\cvprPaperID{5527} 
\def\confName{CVPR}
\def\confYear{2023}

\begin{document}

\title{AeDet: Azimuth-invariant Multi-view 3D Object Detection}

\author{Chengjian Feng
\quad
Zequn Jie
\quad
Yujie Zhong
\quad
Xiangxiang Chu
\quad
Lin Ma\\
\\
Meituan Inc.\\
}
\maketitle

\begin{abstract}
Recent LSS-based multi-view 3D object detection has made tremendous progress, by processing the features in Brid-Eye-View (BEV) via the convolutional detector. However, the typical convolution ignores the radial symmetry of the BEV features and increases the difficulty of the detector optimization. 
To preserve the inherent property of the BEV features and ease the optimization, we propose an azimuth-equivariant convolution (AeConv) and an azimuth-equivariant anchor.
The sampling grid of AeConv is always in the radial direction, thus it can learn azimuth-invariant BEV features.
The proposed anchor enables the detection head to learn predicting azimuth-irrelevant targets.
In addition, we introduce a camera-decoupled virtual depth to unify the depth prediction for the images with different camera intrinsic parameters. 
The resultant detector is dubbed Azimuth-equivariant Detector (AeDet).
Extensive experiments are conducted on nuScenes, and AeDet achieves a 62.0\% NDS, surpassing the recent multi-view 3D object detectors such as PETRv2 and BEVDepth by a large margin.
Project page: \url{https://fcjian.github.io/aedet}.
\end{abstract}
\section{Introduction}
\label{sec:intro}

In the field of autonomous driving, multi-view 3D object detection has been one of the most widely researched problems and received a lot of attention due to its low assembly cost and high efficiency. In the recent literature, such vision-based 3D object detector has made tremendous progress, especially for the Lift-Splat-Shoot (LSS) based methods~\cite{huang2021bevdet, li2022bevdepth}. They first transfer the image features from the image-view to Bird-Eye-View~(BEV), and then process the BEV features via the convolutional backbone and detection head similar to 2D object detection \cite{yin2021center,feng2021tood,feng2021exploring}.

However, the BEV features in multi-view 3D object detection are significantly different from the image features in 2D object detection: (1) the BEV features from the same camera are naturally endowed with radial symmetry; (2) the cameras
have different orientations in BEV, and the BEV features from the different cameras are also approximately with radial symmetry. Simply using the
typical 2D backbone and detection head to perform BEV perception (such as~\cite{huang2021bevdet,li2022bevdepth}) ignores the inherent property of the BEV features and suffers from two limitations discussed as follows:

\begin{figure}[t]
		\centering
		\includegraphics[width=8.1cm]{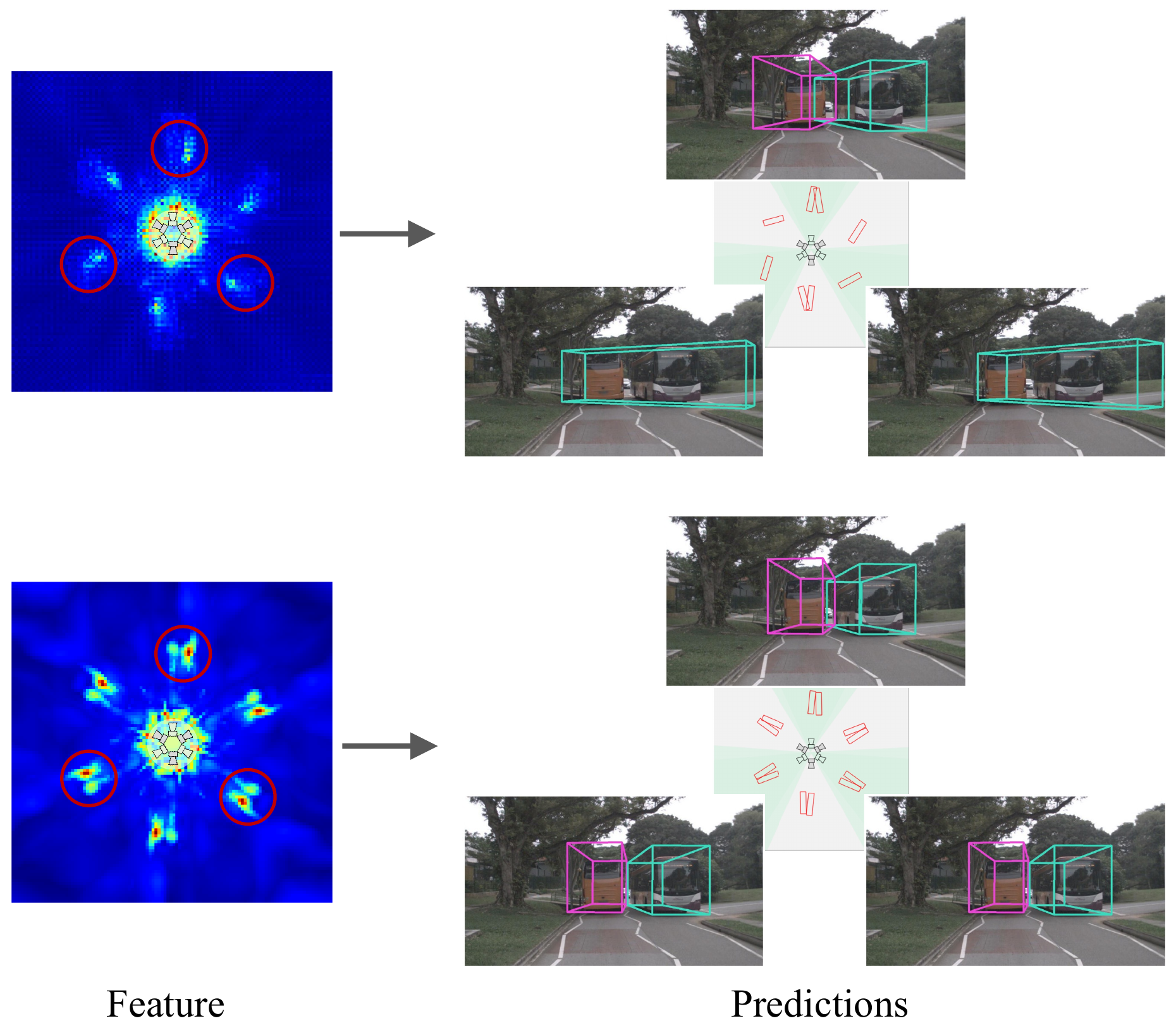}
		\vspace{-1.8mm}
		\caption{Illustration of the BEV features and predictions from the typical LSS-based detector BEVDepth \cite{li2022bevdepth} (top row) and our AeDet (bottom row). The BEV features are the outputs of the BEV backbone. Assume the six cameras capture the same imaging, and the detector takes the same imaging as the input of the six views. BEVDepth generates different features and predictions for the same bus in different azimuths, while AeDet yields almost the same feature and prediction for the same bus in different azimuths.}
		\vspace{-3.5mm}
		\label{demo:intro}
\end{figure}

\begin{figure*}
  \centering
  \begin{subfigure}{0.48\linewidth}
    \centering
    \includegraphics[height=3.8cm]{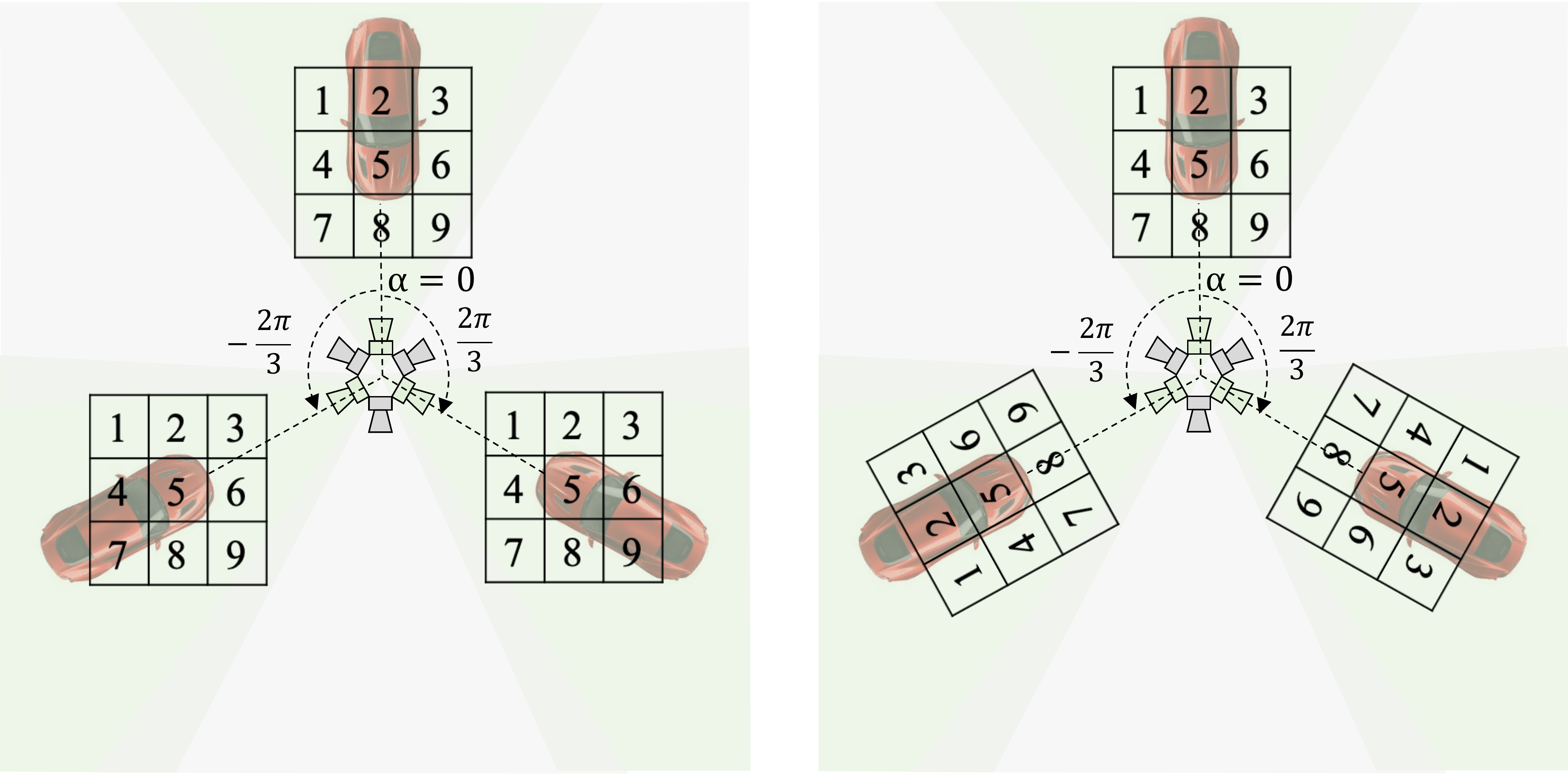}
    \caption{Typical convolution \vs azimuth-equivariant convolution.}
    \label{demo:rotatedconv}
  \end{subfigure}
  \hfill
  \begin{subfigure}{0.48\linewidth}
    \centering
    \includegraphics[height=3.8cm,trim=0 0 0 0,clip]{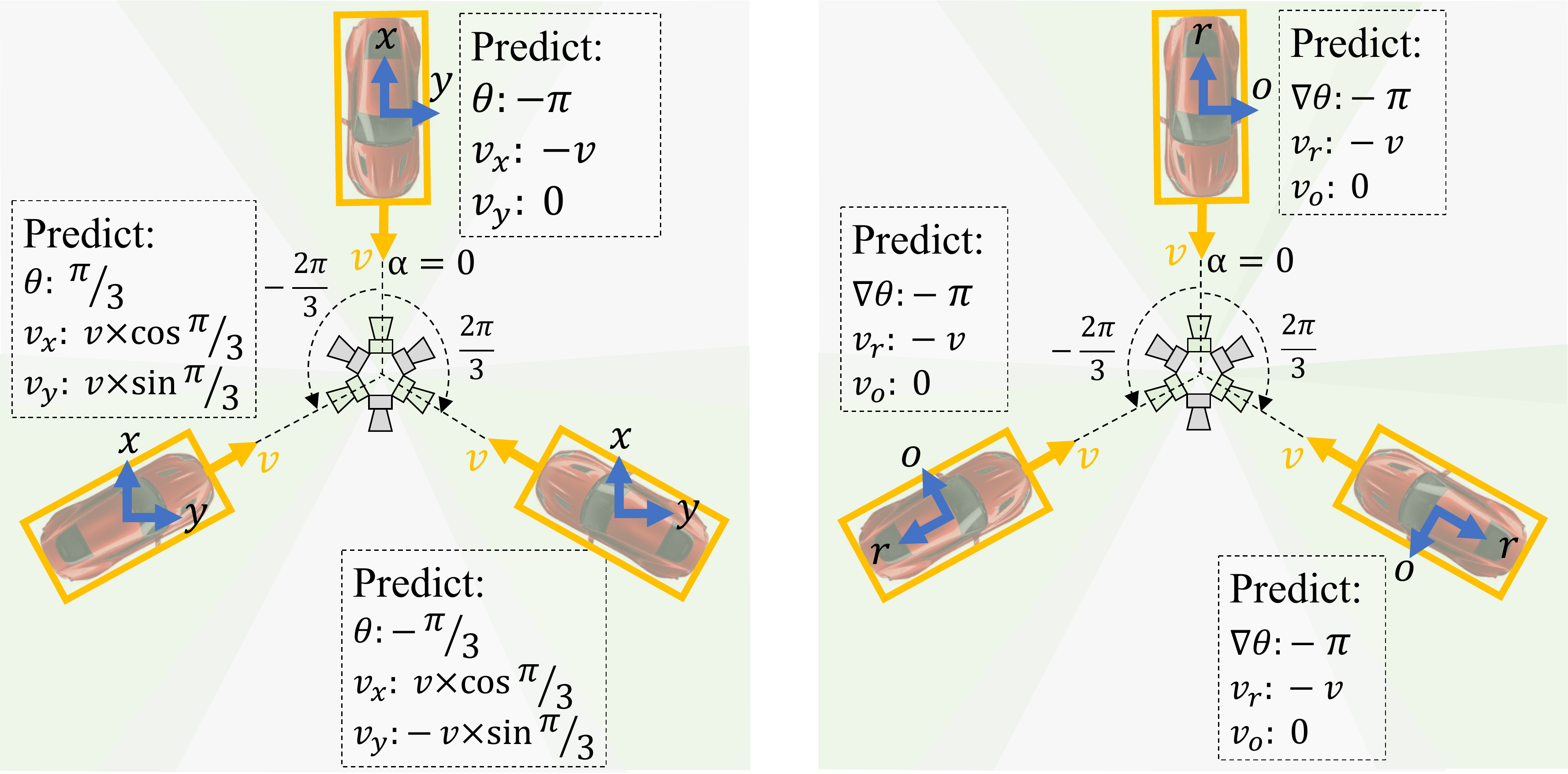}
    \caption{Typical anchor \vs azimuth-equivariant anchor.}
    \label{demo:rotatedanchor}
  \end{subfigure}
  \label{demo:rotatednet}
  \vspace{-1.5mm}
  \caption{Illustration of the azimuth-equivariant convolution and the azimuth-equivariant anchor. Assume the same car precisely faces different-view cameras with the same distance at different timestamps, and the cameras capture the same imaging for the car. These images would be transferred to be the rotation-equivalent BEV features in different azimuths (illustrated by the red cars). The yellow boxes/arrows mean the ground-truth boxes/orientations, and the blue arrows mean the anchor orientation~(`$x$'/`$r$') and its orthogonal direction~(`$y$'/`$o$').}
  \vspace{-3mm}
\end{figure*}

\vspace{1mm}
\noindent \textbf{First, the BEV representation of the same imaging in different azimuths is inconsistent.} The typical convolution shares the kernel weights and adopts the same regular sampling grid at each location of the features. Such design may destroy the radial symmetry of the BEV features, and thus is unfriendly to representation learning. 
To be specific, assume the six cameras capture the same imaging of the object, and the detector takes the same imaging as the input of the six views. 
These images are then transferred to be the rotation-equivalent features in different azimuths from BEV.
However, the sampling grid of the convolution is translation-invariant and samples the inconsistent BEV features of the object in different azimuths (see Figure~\ref{demo:rotatedconv} for detailed demonstration). Consequently, as demonstrated in the `BEV feature' column in Figure~\ref{demo:intro}, the convolution of BEVDepth learns different features in different azimuths, increasing the difficulty of the representation learning.

\vspace{1mm}
\noindent \textbf{Second, the prediction targets of the same imaging in different azimuths are inconsistent.}
The typical detection head predicts the object orientation and velocity along the Cartesian coordinates, which requires the detection head to predict different targets for the same imaging in different azimuths. 
Concretely, assume the different-view cameras capture the same imaging of the object at different moments.
After mapping the imaging from the image to BEV, the object would have different orientations and velocities along the Cartesian coordinates in different azimuths (see Figure~\ref{demo:rotatedanchor} for detailed demonstration). As a result, the detection head is required to predict different targets even for the same imaging in different azimuths, and inevitably increases the difficulty of the predictions.

To address the two limitations, we propose an Azimuth-equivariant Detector~(AeDet) that aims to perform an azimuth-invariant BEV perception by modeling the property of radial symmetry to the network: 
\textbf{(1) Azimuth-equivariant convolution.} In contrast to the typical convolution that uses the same regular sampling grid at each location, we design an Azimuth-equivariant Convolution~(AeConv) to rotate the sampling grid according to the azimuth at each location. AeConv enables the sampling grid equivariant to the azimuth and always in the radial direction of the camera. This allows the convolution to preserve the 
radial symmetry of the BEV features and unify the representation in different azimuths.
\textbf{(2) Azimuth-equivariant anchor.} To unify the prediction targets in different azimuths, we propose an azimuth-equivariant anchor. Specifically, different from the typical anchor~(anchor point or anchor box) defined along the Cartesian coordinates, the azimuth-equivariant anchor is defined along the radial direction and equivariant to the azimuth. We predict both the bounding box and velocity along the new anchor and its orthogonal directions, yielding the same prediction target for the same imaging of the object in different azimuths.
Thus the azimuth-equivariant anchor enables the
detection head to learn predicting azimuth-irrelevant targets.
Notably, \conv~and the azimuth-equivariant anchor can work collaboratively to improve the consistency between the representation learning and predictions, as shown in the `BEV feature' and `Predictions' columns of AeDet in Figure~\ref{demo:intro}.

In addition, we introduce a camera-decoupled virtual depth to improve the depth prediction, and ease the optimization of the depth network. In specific, we decouple the camera's intrinsic parameters from the depth network, enabling the depth network to model the relationship between the image feature and the virtual depth. 
In this way, the depth network only needs to learn to predict a universal virtual depth, regardless of the intrinsic parameters of different cameras.
Finally, we map the virtual depth to the real depth according to the classic camera model.

To summarize, we make the following contributions:
(1) We design an azimuth-equivariant convolution to unify the representation learning in different azimuths, and extract the azimuth-invariant BEV features.
(2) We propose a new azimuth-equivariant anchor to redefine the anchor along the radial direction and unify the prediction targets in different azimuths.
(3) We introduce a camera-decoupled virtual depth to unify the depth prediction for the images captured by different cameras.
(4) We conducted extensive experiments on nuScenes \cite{caesar2020nuscenes}, where our AeDet significantly improves the accuracy of the object orientation (by 5.2\%) and velocity (by 6.6\%). AeDet achieves 62.0\% NDS on the nuScenes \emph{test} set, surpassing the recent multi-view object detectors such as BEVDepth \cite{li2022bevdepth} by a large margin.
\begin{figure*}[t]
		\centering
		\includegraphics[height=6.5cm]{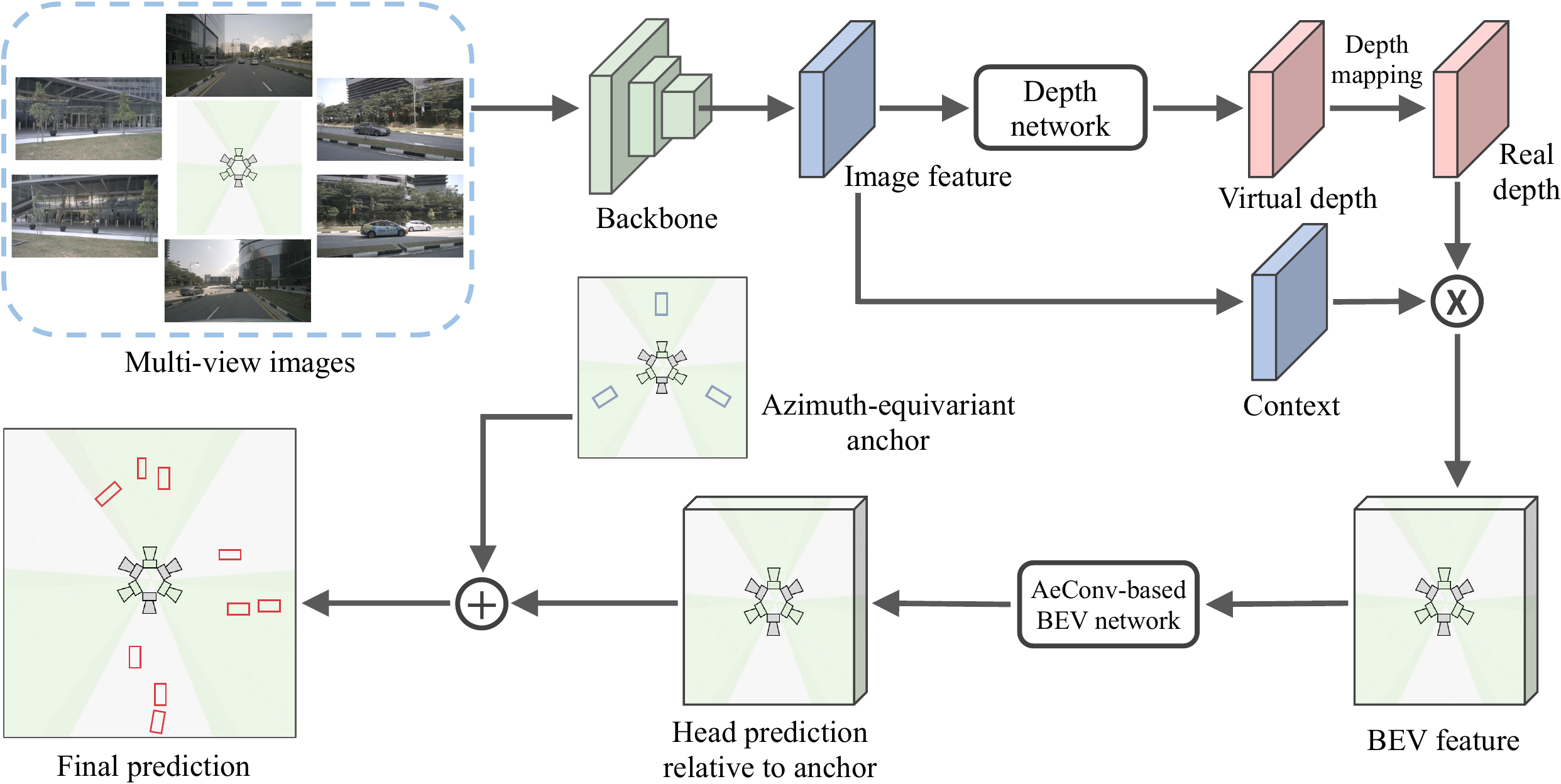}
		\vspace{-1.5mm}
		\caption{Framework of the proposed AeDet. The multi-view images are processed by the image backbone and the view transformer to generate the BEV features with radial symmetry. Then the AeConv-based BEV network further encoders the azimuth-invariant BEV features, and predicts the detection results based on the azimuth-equivariant anchor.}
		\vspace{-2mm}
		\label{demo:structure}
\end{figure*}

\vspace{-1.5mm}
\section{Related Work}
\vspace{-0.5mm}
\noindent \textbf{Single-view 3D object detection.}
Many works attempt to predict the 3D bounding boxes directly from the single-view image. 
For example, 
Ding \etal \cite{ding2020learning} proposes to use the estimated depth map to improve the image representation for 3D object detection. 
Cai \etal \cite{cai2020monocular} inversely projects the 2D structured polygon to a 3D cuboid by providing the object height prior.
Wang \etal \cite{wang2021fcos3d} makes FCOS \cite{tian2019fcos} work in 3D detection by decoupling the 3D targets as 2D and 3D attributes.
Different from these methods, some researchers \cite{wang2019pseudo,wang2019pseudo} predict the 3D bounding boxes based on a pseudo-LiDAR. They convert the image-based depth to pseudo-LiDAR representations and adopt the LiDAR-based detectors to obtain the 3D predictions. Instead of pseudo-LiDAR, Reading \etal \cite{reading2021categorical} projects the image feature to the efficient BEV feature to perform 3D object detection.

\vspace{1mm}
\noindent \textbf{Multi-view 3D object detection.}
Generally speaking, modern 3D object detection frameworks can be divided into LSS-based and query-based methods. LSS-based methods project the image feature from the image-view to BEV using Lift-Splat-Shoot \cite{philion2020lift}, and process the BEV features via the BEV-based detectors~\cite{yin2021center}. Huang \etal \cite{huang2021bevdet} first proposes the LSS-based detector BEVDet for multi-view object detection. Li \etal \cite{li2022bevdepth} enhances the depth prediction by introducing a camera-aware depth network and supervising the network with the depth generated from LiDAR point cloud. In contrast, query-based methods project the object query to the image-view and sample the image features to perform 3D object detection with transformer. Wang \etal \cite{wang2022detr3d} first extends DETR \cite{carion2020end} to DETR3D for multi-view 3D object detection. Liu \etal \cite{liu2022petr} further improves DETR3D with the 3D position-aware representations. Li \etal \cite{li2022bevformer} leverages the deformable transformer to fuse the image features.
Recent works \cite{chen2022polar,jiang2022polarformer} adopt the Polar Parametrization to unify the prediction targets.
Different from these methods, we propose AeDet to improve the LSS-based detection, by learning both the azimuth-invariant BEV representation and the azimuth-irrelevant prediction.

\vspace{1mm}
\noindent \textbf{Single-view depth estimation.}
Single-view depth estimation aims to predict the depth from a single image. 
Saxena \etal \cite{saxena2005learning} collects the monocular images and the corresponding ground-truth depth, and learns the depth with a supervised learning manner.
Garg \etal \cite{garg2016unsupervised} learns the single-view depth via an unsupervised framework from the stereo image pair.
Kuznietsov~\etal \cite{kuznietsov2017semi} trains the network in a semi-supervised way where they use the sparse ground-truth depth for supervision.
Fu \etal \cite{fu2018deep} recasts the depth network learning as an ordinal regression problem and uses a multi-scale network to improve the depth prediction.
In the multi-view 3D detection, Huang \etal \cite{huang2021bevdet} learns the single-view depth with the supervision from the 3D bounding boxes, and Li \etal \cite{li2022bevdepth} improves the depth learning by using the supervision from the LiDAR point cloud.

\section{Methodology}
\label{sec:method}
In this paper, we aim to make an azimuth-invariant Bird-Eye-View~(BEV) perception for multi-view 3D object detection, and thus unify the representation learning and predictions in different azimuths from BEV and boost the detection performance, as illustrated in Figure~\ref{demo:structure}. This section is organized as follows, we start by introducing the common paradigm of the LSS-based multi-view 3D object detection. In Section~\ref{subset:rotatednetwork}, we propose an Azimuth-equivariant Convolution~(AeConv) to extract the azimuth-invariant BEV representation. In Section~\ref{subset:rotatedanchor}, we further design a new azimuth-equivariant anchor to unify the prediction targets in different azimuths. Additionally, we introduce a camera-decoupled virtual depth to unify the depth prediction for the multi-camera images and ease the prediction in Section~\ref{subset:depthnet}.

\subsection{LSS-based Multi-view 3D Object Detection}
\label{subsec:bevdepth}
Generally speaking, a popular LSS-based multi-view 3D object detector, for example, BEVDepth~\cite{li2022bevdepth}, is consisted of an image-view encoder, a view transformer, a BEV encoder and a detection head.

\vspace{1mm}
\noindent \textbf{Image-view encoder.} It is a typical 2D backbone such as ResNet \cite{he2016deep}, with the aim to extract the high-level features of the multi-view images.

\vspace{1mm}
\noindent \textbf{View transformer.} It first predicts the depth based on the image features, and then transforms the image features from image-view to BEV according to the predicted depth. BEVDet~\cite{huang2021bevdet} predicts the depth using a camera-agnostic depth network. BEVDepth~\cite{li2022bevdepth} improves the depth network by introducing the camera's intrinsic parameters into the features with a camera-aware attention module.

\vspace{1mm}
\noindent \textbf{BEV encoder.} It adopts a typical convolutional encoder similar to the image-view 2D backbone to further encode the BEV features.

\vspace{1mm}
\noindent \textbf{Detection head.} It usually leverages the head structure designed in LiDAR point-cloud based 3D object detection, for example, CenterPoint~\cite{yin2021center}. The detection head of CenterPoint~(the first stage) adopts an anchor-free structure, which predicts the center offset, size, orientation and velocity of the object at each location.

\subsubsection{Discussion on Limitations} 
The BEV features from the same camera are naturally endowed with radial symmetry. Furthermore, the cameras have different orientations in BEV, and naively using the typical 2D backbone and the anchor-free detection head to perform BEV perception would increase the difficulty of the detector optimization in two aspects: 

\noindent (1) the convolution applies the same sampling grid $\mathcal{R}$ at each location of the features, \eg $\mathcal{R}=\{(-1, -1), (-1, 0), \ldots, (0,1), (1, 1)\}$ for a $3 \times 3$ convolutional kernel with dilation $1$. It would sample and produce different BEV representations even for the same imaging in different azimuths, as illustrated in Figure~\ref{demo:rotatedconv}; 

\noindent (2) the anchor-free detection head directly predicts the object orientation and velocity along the Cartesian coordinates, which would require the detection head to predict different targets for the same imaging in different azimuths, as illustrated in Figure~\ref{demo:rotatedanchor}. 

In the following sections, we propose an Azimuth-equivariant Detector~(AeDet) to learn the azimuth-invariant BEV representation and unify the prediction targets in different azimuths, and thus ease the optimization of the multi-view 3D object detection.

\subsection{Learning Azimuth-invariant BEV Feature}
\label{subset:rotatednetwork}
To unify the representation learning in different azimuths, we propose an azimuth-equivariant convolution, termed \conv, to extract the representation along the radial direction of the camera.

\vspace{1mm}
\noindent \textbf{\conv.} To introduce \conv, we consider a simple case with only one single-view camera. Firstly, we define an azimuth system based on the camera, in which we let the focal point of the camera as the center point of the system, the clockwise from BEV as the ascending direction of the azimuth $\alpha$, and the ego-direction as the reference direction with an azimuth of 0. According to the azimuth at each location, we rotate the regular sampling grid $\mathcal{R}$ of the typical convolution to a rotated version $\mathcal{R}^{rot}$ by:
\begin{equation}
\boldsymbol{p}_i^{rot}
 = 
\left[
\begin{array}{cc}
    \cos(\alpha) & \sin(\alpha) \\
    -\sin(\alpha) & \cos(\alpha)
\end{array}
\right]
\boldsymbol{p}_i,
\end{equation}
where $\boldsymbol{p}_{i} \in \mathcal{R}$, $\mathcal{R}^{rot} = \{\boldsymbol{p}_i^{rot} | i\in[1, 2, \ldots, \rm K]\}$, and $\rm K$ is the number of the sampling locations. 
Then we adopt the convolution operation based on the new sampling grid:
\begin{equation}
\boldsymbol{y}=\sum_{\boldsymbol{p}_i, \boldsymbol{p}_i^{rot}}\boldsymbol{w}(\boldsymbol{p}_i)\cdot \boldsymbol{x}(\boldsymbol{p}_i^{rot}),
\label{eq.standard_conv}
\end{equation}
where $\boldsymbol{x}$ and $\boldsymbol{w}$ denote the feature map and the convolutional kernel, respectively. $\boldsymbol{x}(\boldsymbol{p}_i^{rot})$ is implemented by bi-linear interpolation. 
\emph{Note that the rotated sampling grid is always in the radial direction of the camera} (illustrated in Figure \ref{demo:rotatedconv}). Therefore, \conv~could sample and learn the same BEV feature for the same imaging even in different azimuths, \ie the azimuth-invariant representation.

\vspace{1mm}
\noindent \textbf{BEV network based on \conv.} In the multi-view autonomous driving system, the positions of the cameras are different, resulting in non-uniform azimuth system for different cameras. To align the azimuth system and apply a unified AeConv in multi-view BEV features, we propose to use the average center of the cameras as an approximate center point of the azimuth system. The ascending direction and reference direction of the azimuth are the same as the ones defined above.
In this work, we build an AeConv-based BEV network by replacing the typical convolution with the proposed \conv.
It enables to extract the azimuth-invariant BEV features along the radial direction, largely easing the optimization of the BEV network.

\begin{figure}[t]
		\centering
		\includegraphics[height=4.4cm]{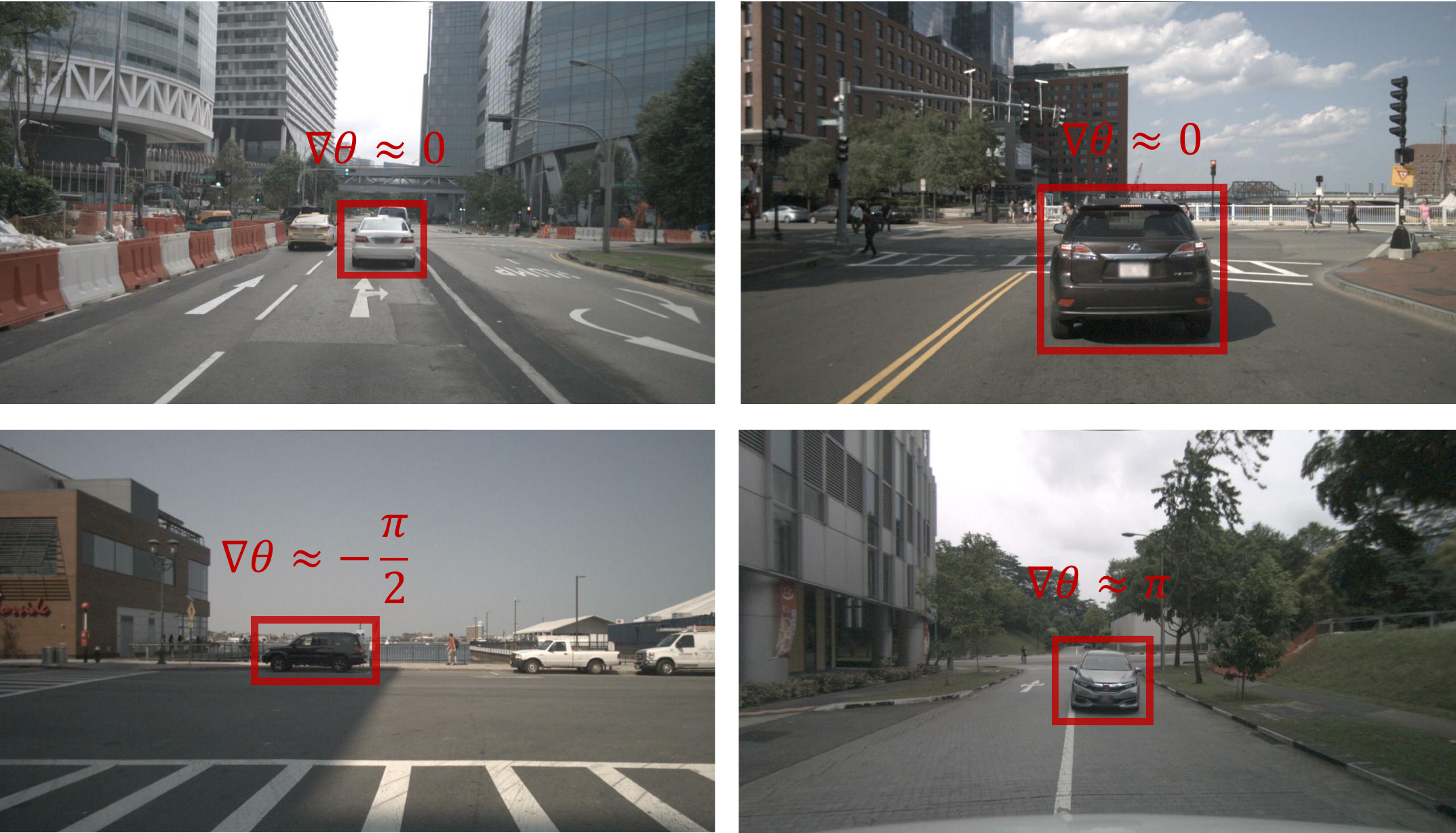}
		\vspace{-1mm}
		\caption{Illustration of the relative angles of the car orientation $w.t.r$ the radial direction of the camera. We could easily predict these relative angles from the images.}
		\vspace{-2mm}
		\label{demo:car_images}
\end{figure}

\subsection{Azimuth-equivariant Anchor}
\label{subset:rotatedanchor}
Till here, we have extracted the azimuth-invariant representation. In this section, we further introduce a new azimuth-equivariant anchor to enable the detection head to make a unified prediction in different azimuths. The detection head based on the azimuth-equivariant anchor predicts the detection results different from the previous methods in two aspects. First, it predicts the relative angle of the object orientation $w.r.t$ the azimuth at each location, instead of the absolute object orientation. Second, it predicts both the center offset and velocity along the radial direction and its orthogonal direction. In the following, we first discuss the implicit anchor existing in the anchor-free methods, and then elaborate on the proposed azimuth-equivariant anchor.

\vspace{1mm}
\noindent \textbf{Implicit anchor.}
The anchor-free detection head is the most commonly used head structure in LSS-based 3D object detectors such as BEVDet and BEVDepth. 
It actually could be seen as a specific case of the anchor-based detection head with an implicit anchor, namely, an anchor box with zero length, width and height.

To be specific, the anchor-based detection head densely defines the anchor $(x, y, z, l, w, h, \theta)$ at each location, where $(x, y, z)$, $(l, w, h)$ and $\theta$ are the location, size and orientation of the anchor respectively. 
The anchor-free detection head follows a similar paradigm, but sets $l = w = h = \theta = 0$, which can be viewed as having implicit anchors.

\vspace{1mm}
\noindent \textbf{Azimuth-equivariant anchor.} 
Due to the azimuth-irrelevant design, the existing anchor-free detection head is required to predict different targets even for the same imaging in different azimuths. This inevitably increases the difficulty of the predictions.
To ease the prediction, we design an azimuth-equivariant anchor to enable the detection head to predict the same targets for the same imaging, \ie being azimuth-independent. Essentially, the detection head can therefore predict the relative angle~(shown in Figure~\ref{demo:car_images}) and the velocity/center offset along the consistent directions~(shown in Figure~\ref{demo:rotatedanchor}).

Specifically, similar to \conv, we define an azimuth-equivariant anchor $(x, y, z, l, w, h, \alpha)$ according to the azimuth at each location.
Then we could compute the box residual (\ie prediction target) between the azimuth-equivariant anchors and the marching ground-truth boxes. The orientation residual is the relative angle of the object orientation $w.r.t$ the azimuth:
\begin{equation}
\nabla \theta = \hat{\theta} - \alpha,
\end{equation}
where $\hat{\theta}$ denotes the object orientation. In contrast to the existing methods~\cite{huang2021bevdet,li2022bevdepth} that predict the center offset $(\nabla x, \nabla y)$ and the velocity $(v_{x}, v_{y})$ along the Cartesian coordinates, we predict them along the anchor orientation and its orthogonal direction. The new center offset $(\nabla r, \nabla o)$ and the velocity $(v_{r}, v_{o})$ can be computed by:
\begin{equation}
\left[
\begin{array}{cc}
    \nabla r \\
    \nabla o
\end{array}
\right]
 = 
\left[
\begin{array}{cc}
    \cos(\alpha) & \sin(\alpha) \\
    -\sin(\alpha) & \cos(\alpha)
\end{array}
\right]
\left[
\begin{array}{cc}
    \nabla x \\
    \nabla y
\end{array}
\right],
\end{equation}
\begin{equation}
\left[
\begin{array}{cc}
    v_{r} \\
    v_{o}
\end{array}
\right]
 = 
\left[
\begin{array}{cc}
    \cos(\alpha) & \sin(\alpha) \\
    -\sin(\alpha) & \cos(\alpha)
\end{array}
\right]
\left[
\begin{array}{cc}
    v_{x} \\
    v_{y}
\end{array}
\right].
\end{equation}
\emph{Note that the orientation of the azimuth-equivariant anchor is always in the radial direction of the camera.} Therefore, the detection head could predict the orientation, center offset and velocity of the object based on the azimuth-equivariant direction, yielding the azimuth-irrelevant prediction targets, as illustrated in Figure~\ref{demo:rotatedanchor}.

\begin{table*}
\vspace{-2mm}
\centering
\begin{tabular}{cc@{\hspace{1.7\tabcolsep}}c@{\hspace{1.7\tabcolsep}}cccccccc}
\toprule
\textbf{AeNet} & \textbf{CDN} & \textbf{Params (M)} & \textbf{FLOPs (G)} & \textbf{mAP}$\uparrow$  & \textbf{mATE}$\downarrow$ & \textbf{mASE}$\downarrow$  & \textbf{mAOE}$\downarrow$ & \textbf{mAVE}$\downarrow$ & \textbf{mAAE}$\downarrow$ & \textbf{NDS}$\uparrow$ \\
\cmidrule(r){1-1}
\cmidrule(r){2-2}
\cmidrule(r){3-4}
\cmidrule(r){5-10}
\cmidrule(r){11-11}
& & 74.12 & 459.22 & 0.330 & 0.699 & 0.281 & 0.545 & 0.493 & 0.212 & 0.442 \\
\cmark & & 74.12 & 467.04 & 0.348 & 0.675 & 0.275 & 0.509 & 0.445 & 0.218 & 0.462 \\
& \cmark & 74.02 & 465.28 & 0.345 & 0.682 & 0.278 & 0.512 & 0.489 & 0.225 & 0.454 \\
\cmark & \cmark & 74.02 & 473.10 & 0.358 & 0.655 & 0.273 & 0.493 & 0.427 & 0.216 & 0.473 \\
\bottomrule
\end{tabular}
\vspace{-2mm}
\caption{Ablation study of the components in AeDet. AeNet denotes AeConv and the azimuth-equivariant anchor, and CDN denotes the camera-decoupled depth network.}
\vspace{-5mm}
\label{tab:ablation}
\end{table*}

\subsection{Camera-decoupled Virtual Depth}
\label{subset:depthnet}

\begin{figure}[t]
		\centering
		\includegraphics[width=8cm]{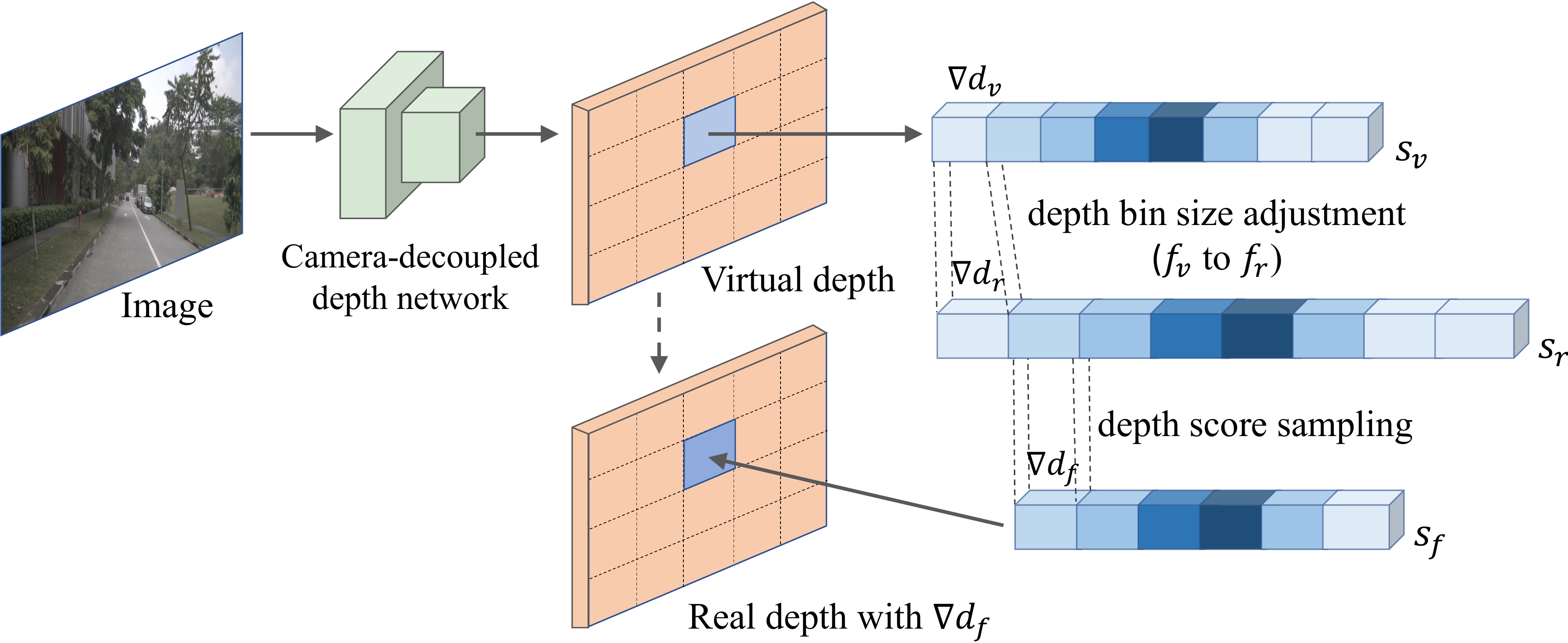}
		\vspace{-2mm}
		\caption{Illustration of the camera-decoupled virtual depth and depth mapping.}
		\vspace{-2mm}
		\label{demo:depthnet}
\end{figure}

The depth plays an important role in rendering a precise BEV feature. Recent BEVDepth improves the depth network by introducing the camera's intrinsic parameters into the features with a camera-aware attention module. However, naively introducing the camera's intrinsic parameters into the features would require the depth network to model the relationship among the image feature, the focal length of the cameras and the depth, and thus increase the difficulty of the depth prediction. To reduce the burden of the depth network, we decouple the camera from the depth network. The depth network can now focus on modeling the relationship between image feature and depth, regardless of the intrinsic parameters of the camera (illustrated in Figure~\ref{demo:depthnet}).

The proposed method contains two steps: (1) we first leverage a camera-decoupled depth network to predict a virtual depth based on a virtual focal length; (2) we then map the virtual depth to the real depth according to the focal length of the camera. We elaborate on the two steps in the following:

\vspace{1mm}
\noindent \textbf{Virtual depth prediction.} We leverage a camera-decoupled depth network to predict a virtual depth that is independent of the intrinsic parameters of the camera. For simplification, we use the same structure of the depth network as BEVDepth, but remove its camera-aware attention module. 
To decouple the camera's intrinsic parameters from the depth network, we assume all the images are captured by the cameras with the same virtual focal length $f_{v}$. 
We then densely predict the virtual depth using only the image feature, through a classification manner, where the classification scores $\boldsymbol{s}_{v} \in \mathbb{R}^{\rm M}$ are predicted for $\rm M$ discretized depth bins, and the $\rm M$ discretized depth bins represent the depth range $[0, d_{v}]$ uniformly. Thus the virtual bin size is:
\begin{equation}
\nabla d_{v} = \frac{d_{v}}{\rm M}.
\end{equation}
In this assumption, the depth network is only required to model the relationship between the image feature and the virtual depth. 
The virtual depth could be transferred to the real depth according to the classic camera model.
The depth is relevant to the object scale. However, the context feature is expected to be scale-invariant to improve the recognition capability of the model. To alleviate the conflict between depth and context, we adopt a deformable convolution in the context branch.

\vspace{1mm}
\noindent \textbf{Depth mapping.}
Note that we could easily transfer the virtual bin size to the real bin size according to the virtual focal length and the real focal lengths:
\begin{equation}
\nabla d_{r} = \frac{f_{r}}{f_{v}} \nabla d_{v},
\end{equation}
\begin{equation}
f_{r} = \sqrt{\frac{f_{x}^{2} + f_{y}^{2}}{2}},
\end{equation}
where $f_{x}$ and $f_{y}$ are the focal lengths of the camera. However, $\nabla d_{r}$ varies with the focal lengths, which would lead to the different densities of the multi-camera BEV features after view transformation. To solve this problem, we map the variable bin size to a fixed one. Assuming the fixed depth bin size is defined as:
\begin{equation}
\nabla d_{f} = \frac{d_{f1} - d_{f2}}{\rm N},
\end{equation}
where $\rm N$ discretized depth bins represent the depth range $[d_{f1}, d_{f2}]$ uniformly. Then our goal is to predict the classification scores $\boldsymbol{s}_{f} \in \mathbb{R}^{\rm N}$ for the $\rm N$ discretized depth bins. $\boldsymbol{s}_{f}$ could be transferred from $s_{v}$ by:
\begin{equation}
\boldsymbol{s}_{f}[i] = \boldsymbol{s}_{v}[\frac{d_{f1} + i \times \nabla d_{f}}{\nabla d_{r}}],
\end{equation}
where $i$ denotes the index of the $\rm N$ fixed depth bins, and $\boldsymbol{s}_{v}[\frac{d_{f1} + i \times \nabla d_{f}}{\nabla d_{r}}]$ is implemented by linear interpolation.

\begin{table}
\centering
\begin{tabular}{lccccc}
\toprule
\textbf{Method}          & \textbf{mAP}$\uparrow$  & \textbf{mAOE}$\downarrow$ & \textbf{mAVE}$\downarrow$ & \textbf{NDS}$\uparrow$ \\
\cmidrule(r){1-1}
\cmidrule(r){2-4}
\cmidrule(r){5-5}
BEVDepth & 0.315 & 0.621 & 1.042 & 0.367 \\
AeDet & 0.331 & 0.563 & 0.895 & 0.397 \\
\bottomrule
\end{tabular}
\vspace{-2mm}
\caption{Comparison between BEVDepth and AeDet in the single-frame scheme.}
\vspace{-3mm}
\label{tab:aedet}
\end{table}

\begin{figure*}[t]
		\centering
		\vspace{-4mm}
		\includegraphics[width=17.1cm]{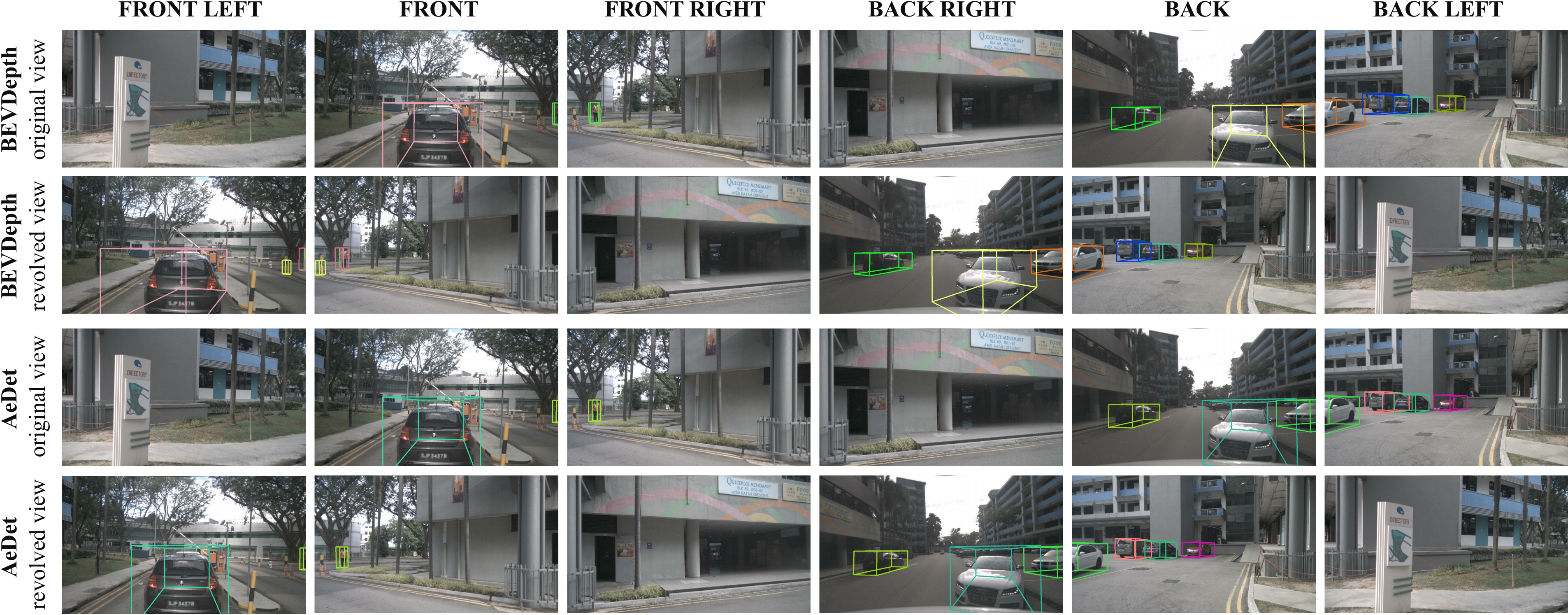}
		\vspace{-2mm}
		\caption{Illustration of detection results from BEVDepth and AeDet in original view and revolved view.}
		\vspace{-2mm}
		\label{demo:revolving}
\end{figure*}

\begin{table}
\centering
\begin{tabular}{cccc}
\toprule
\textbf{M}          & \textbf{mAP}$\uparrow$  & \textbf{mATE}$\downarrow$ & \textbf{NDS}$\uparrow$ \\
\cmidrule(r){1-1}
\cmidrule(r){2-3}
\cmidrule(r){4-4}
150 & 0.353 & 0.665 & 0.465 \\
180 & 0.358 & 0.655 & 0.473 \\
210 & 0.354 & 0.654 & 0.469 \\
\bottomrule
\end{tabular}
\vspace{-2mm}
\caption{Analysis of different numbers of virtual depth bins.}
\vspace{-6mm}
\label{tab:depthbins}
\end{table}

\begin{table*}
\centering
\begin{tabular}{l@{\hspace{1.5\tabcolsep}}c@{\hspace{1.5\tabcolsep}}cccccccc}
\toprule
\textbf{Method} & \textbf{Original view} & \textbf{Revolved view} & \textbf{mAP}$\uparrow$  & \textbf{mATE}$\downarrow$ & \textbf{mASE}$\downarrow$  & \textbf{mAOE}$\downarrow$ & \textbf{mAVE}$\downarrow$ & \textbf{mAAE}$\downarrow$ & \textbf{NDS}$\uparrow$ \\
\cmidrule(r){1-1}
\cmidrule(r){2-3}
\cmidrule(r){4-9}
\cmidrule(r){10-10}
BEVDepth & \cmark & & 0.330 & 0.699 & 0.281 & 0.545 & 0.493 & 0.212 & 0.442 \\
& & \cmark &  0.322 & 0.730 & 0.282 & 0.701 & 0.720 & 0.224 & 0.396 \\
\cmidrule(r){1-1}
\cmidrule(r){2-3}
\cmidrule(r){4-9}
\cmidrule(r){10-10}
AeDet & \cmark &  & 0.358 & 0.655 & 0.273 & 0.493 & 0.427 & 0.216 & 0.473 \\
& & \cmark & 0.357 & 0.657 & 0.272 & 0.499 & 0.426 & 0.221 & 0.471 \\
\bottomrule
\end{tabular}
\vspace{-2mm}
\caption{Performance of BEVDepth and AeDet in original view and revolved view.}
\vspace{-6mm}
\label{tab:revolving}
\end{table*}

\begin{table*}
\vspace{-1mm}
\centering
\resizebox{\textwidth}{!}{
\begin{tabular}{lccc@{\hspace{1.0\tabcolsep}}c@{\hspace{1.0\tabcolsep}}c@{\hspace{1.0\tabcolsep}}c@{\hspace{1.0\tabcolsep}}c@{\hspace{1.0\tabcolsep}}c@{\hspace{1.0\tabcolsep}}c} 
\toprule
\textbf{Method} & \textbf{Backbone} & \textbf{Image size} & \textbf{mAP}$\uparrow$  & \textbf{mATE}$\downarrow$ & \textbf{mASE}$\downarrow$   &\textbf{mAOE}$\downarrow$   &\textbf{mAVE}$\downarrow$   &\textbf{mAAE}$\downarrow$  &\textbf{NDS}$\uparrow$  \\
\cmidrule(r){1-1}
\cmidrule(r){2-2}
\cmidrule(r){3-3}
\cmidrule(r){4-9}
\cmidrule(r){10-10}
BEVDet \cite{huang2021bevdet} & ResNet50 & 256$\times$704                                      &  0.298 & 0.725 & 0.279 & 0.589 & 0.860 & 0.245 & 0.379 \\ 
PETR \cite{liu2022petr} & ResNet50 & 384$\times$1056                                       &  0.313 & 0.768 & 0.278 & 0.564 & 0.923 & 0.225 & 0.381 \\ 
BEVDet4D \cite{huang2021bevdet} & ResNet50 & 256$\times$704                                    &  0.322 & 0.703 & 0.278 & 0.495 & 0.354 & 0.206 & 0.457 \\ 
BEVDepth \cite{li2022bevdepth} & ResNet50 & 256$\times$704                                    &  0.351 & 0.639 & \textbf{0.267} & 0.479 & 0.428 & 0.198 & 0.475 \\ 
\textbf{AeDet}~(ours) & ResNet50 & 256$\times$704        & \textbf{0.387} & \textbf{0.598} & 0.276 & \textbf{0.461} & \textbf{0.392} & \textbf{0.196} & \textbf{0.501} \\
\cmidrule(r){1-1}
\cmidrule(r){2-2}
\cmidrule(r){3-3}
\cmidrule(r){4-9}
\cmidrule(r){10-10}
FCOS3D \cite{wang2021fcos3d} & ResNet101-DCN & 900$\times$1600                                &  0.295 & 0.806 & 0.268 & 0.511 & 1.131 & \textbf{0.170} & 0.372 \\ 
PolarDETR-T \cite{chen2022polar} & ResNet101-DCN & 900$\times$1600                  &  0.383 & 0.707 & 0.269 & \textbf{0.344} & 0.518 & 0.196 & 0.488 \\ 
UVTR \cite{li2022unifying} & ResNet101-DCN &
900$\times$1600                         &  0.379 & 0.731 & 0.267 & 0.350 & 0.510 & 0.200 & 0.483 \\ 
PolarFormer \cite{jiang2022polarformer} & ResNet101-DCN & 900$\times$1600                  &  0.432 & 0.648 & 0.270 & 0.348 & 0.409 & 0.201 & 0.528 \\ 
DETR3D \cite{wang2022detr3d} & ResNet101-DCN & 900$\times$1600                    & 0.349 & 0.716 & 0.268 & 0.379 & 0.842 & 0.200 & 0.434 \\ 
BEVFormer \cite{li2022bevformer} & ResNet101-DCN & 900$\times$1600                    &  0.416 & 0.673 & 0.274 & 0.372 & 0.394 & 0.198 & 0.517 \\ 
PETR \cite{liu2022petr} & ResNet101 & 512$\times$1408                                 & 0.357 & 0.710 & 0.270 & 0.490 & 0.885 & 0.224 & 0.421 \\ 
BEVDepth \cite{li2022bevdepth} & ResNet101 & 512$\times$1408                             & 0.412 & 0.565 & 0.266 & 0.358 & 0.331 & 0.190 & 0.535 \\ 
\textbf{AeDet}~(ours) & ResNet101& 512$\times$1408  & \textbf{0.449} & \textbf{0.501} & \textbf{0.262} & 0.347 & \textbf{0.330} & 0.194 & \textbf{0.561}\\
\cmidrule(r){1-1}
\cmidrule(r){2-2}
\cmidrule(r){3-3}
\cmidrule(r){4-9}
\cmidrule(r){10-10}
BEVDepth \cite{li2022bevdepth} & ConvNeXt-B & 512$\times$1408                             & 0.462 &  0.540 & \textbf{0.254} &  0.353 & 0.379 & 0.200  & 0.558 \\ 
\textbf{AeDet}~(ours) & ConvNeXt-B & 512$\times$1408  & \textbf{0.483} & \textbf{0.494} & 0.261 & \textbf{0.324} & \textbf{0.337} & \textbf{0.195} & \textbf{0.581}\\
\bottomrule
\end{tabular}}
\vspace{-2.5mm}
\caption{Comparison on the nuScenes \emph{val} set.}
\vspace{-2.5mm}
\label{tab:val}
\end{table*}

\begin{table*}
\centering
\resizebox{\textwidth}{!}{
\begin{tabular}{lccc@{\hspace{1.0\tabcolsep}}c@{\hspace{1.0\tabcolsep}}c@{\hspace{1.0\tabcolsep}}c@{\hspace{1.0\tabcolsep}}c@{\hspace{1.0\tabcolsep}}c@{\hspace{1.0\tabcolsep}}c} 
\toprule
\textbf{Method} & \textbf{Backbone} & \textbf{Image size} & \textbf{mAP}$\uparrow$  & \textbf{mATE}$\downarrow$ & \textbf{mASE}$\downarrow$   &\textbf{mAOE}$\downarrow$   &\textbf{mAVE}$\downarrow$   &\textbf{mAAE}$\downarrow$  &\textbf{NDS}$\uparrow$  \\
\cmidrule(r){1-1}
\cmidrule(r){2-2}
\cmidrule(r){3-3}
\cmidrule(r){4-9}
\cmidrule(r){10-10}
DETR3D \cite{wang2022detr3d}         & V2-99      & 900$\times$1600  & 0.412 & 0.641 & 0.255 & 0.394 & 0.845 & 0.133 & 0.479 \\
UVTR \cite{li2022unifying}          & V2-99      & 900$\times$1600  & 0.472 & 0.577 & 0.253 & 0.391 & 0.508 & 0.123 & 0.551 \\
BEVFormer \cite{li2022bevformer}     & V2-99      & 900$\times$1600  & 0.481 & 0.582 & 0.256 & 0.375 & 0.378 & 0.126 & 0.569 \\
BEVDet4D \cite{huang2021bevdet}      & Swin-B     & 900$\times$1600  & 0.451 & 0.511 & \textbf{0.241} & 0.386 & 0.301 & 0.121 & 0.569 \\
PolarFormer \cite{jiang2022polarformer}   & V2-99      & 900$\times$1600  & 0.493 & 0.556 & 0.256 & 0.364 & 0.439 & 0.127 & 0.572 \\
PETRv2 \cite{liu2022petrv2}         & V2-99  & 640$\times$1600  & 0.490 & 0.561 & 0.243 & 0.361 & 0.343 & \textbf{0.120} & 0.582 \\
BEVDepth~\cite{li2022bevdepth} & V2-99 & 640$\times$1600  & 0.503 & 0.445 & 0.245 & 0.378   & 0.320 & 0.126 & 0.600  \\
BEVDepth~\cite{li2022bevdepth}$^{\dag}$ & ConvNeXt-B & 640$\times$1600  & 0.520 & 0.445 & 0.243 & 0.352   & 0.347 & 0.127 & 0.609  \\
\textbf{AeDet}~(ours) & ConvNeXt-B & 640$\times$1600  & \textbf{0.531} & \textbf{0.439} & 0.247 & \textbf{0.344} & \textbf{0.292} & 0.130 & \textbf{0.620} \\
\bottomrule
\end{tabular}}
\vspace{-2.5mm}
\caption{Comparison on the nuScenes \emph{test} set. $^{\dag}$ denotes data augmentation that randomly samples time intervals in previous frames.}
\vspace{-5mm}
\label{tab:test}
\end{table*}

\vspace{-1mm}
\section{Experiment}
\vspace{-1mm}
\noindent \textbf{Dataset \& Evaluation Metrics.}
All experiments are implemented on the nuScenes benchmark \cite{caesar2020nuscenes}, which includes 1000 scenes with images from six cameras.
Following the standard practice \cite{carion2020end,huang2021bevdet,li2022bevdepth}, we split the dataset into 700, 150, and 150 scenes for training, validation and testing.
The detection performance is measured by the official evaluation metrics: mean Average Precision (mAP), mean Average Translation Error (mATE), mean
Average Scale Error (mASE), mean Average Orientation Error (mAOE), mean Average Velocity Error(mAVE), mean Average Attribute Error (mAAE) and nuScenes Detection Score (NDS).

\vspace{1mm}
\noindent \textbf{Implementation details.}
We use the most recent multi-view 3D object detector BEVDepth \cite{li2022bevdepth} as our baseline. We conduct the experiment with different image backbones including ResNet-50 \cite{he2016deep}, ResNet-101 \cite{he2016deep} and ConvNeXt-B \cite{liu2022convnet}. 
The deconvolution is replaced with the up-sampling layer and convolution in the BEV network. 
Unless specified, we train the model with a total batchsize of 64 for 24 epochs, where AdamW \cite{loshchilov2017decoupled} is adopted as the optimizer with the learning rate of 2e-4 and the EMA technique. Following BEVDepth \cite{li2022bevdepth}, both image and BEV data augmentations are used during training, and $d_{f1}$, $d_{f2}$ and $\nabla d_{f}$ are set to 2m, 54m and 0.5m. 
The mapped real depth range must include the fixed depth range $[2, 54]$m. 
To achieve that, we set $d_{v}$ and $f_{v}$ to 54m (\ie $d_{f2}$) and 800 pixels (\ie less than the minimum focal length of the nuScenes cameras).

\subsection{Ablation Study}
\vspace{-1mm}
For an ablation study, we use the ResNet-50 backbone with the input of 2 key frames and 256$\times$704 resolution unless specified. The performances are reported on the \emph{val} set.

\vspace{1mm}
\noindent \textbf{Azimuth-equivariant network.}
To demonstrate the effectiveness of both AeConv and the azimuth-equivariant anchor, we apply them to BEVDepth. As shown in Table~\ref{tab:ablation}, AeNet (AeConv + the azimuth-equivariant anchor) improves the NDS from 44.2\% to 46.2\%, with the same number of parameters and only +1.7\% FLOPs. To be more specific,  the improvements in mAOE and mAVE are significant: mAOE and mAVE are improved by 3.6\% and 4.8\%, respectively.
This validates the effectiveness of the design of AeConv and the azimuth-equivariant anchor, and demonstrates that they can work efficiently to obtain high performance, especially for the object orientation and velocity.

\vspace{1mm}
\noindent \textbf{Camera-decoupled virtual depth.}
Here, we conduct the ablation study on the proposed camera-decoupled virtual depth.
We first use the camera-decoupled depth network to predict the virtual depth, and then compute the real depth through the depth mapping. 
As shown in Table~\ref{tab:ablation}, the camera-decoupled depth network improves the mAP and NDS by 1.5\% and 1.2\%, comparing to BEVDepth. Specifically, the most improvement is in mATE (69.9\% $\rightarrow$68.2\%). This shows that the proposed virtual depth and depth mapping enhance the accuracy of the depth prediction, by unifying the depth prediction of the multi-camera images.

\vspace{1mm}
\noindent \textbf{AeDet.}
We evaluate the performance of the complete AeDet (\ie AeNet + CDN). As shown in Table~\ref{tab:ablation}, AeDet achieves 35.8\% mAP and 47.3\% NDS, surpassing BEVDepth by 2.8\% mAP and 3.1\% NDS. To be more specific, the mAOE and mAVE are improved by 5.2\% and 6.6\%. We also evaluate the performance of the single-frame AeDet, as shown in Table~\ref{tab:aedet}. Compared with BEVDepth, AeDet brings an improvement of 3.0\% NDS in single-frame scheme. Note that estimating the velocity of the object is difficult in the single-view scheme. By unifying the prediction of velocity, AeDet improves the mAVE by 14.7\%.

\vspace{1mm}
\noindent \textbf{The number of virtual depth bins.}
We investigate the performance using different $\rm M$ which denotes the number of the discretized virtual depth bins. Through a coarse search shown in Table~\ref{tab:depthbins}, we find that the performance is insensitive to $\rm M$. We adopt $\rm M=180$ for our experiments.

\vspace{-1mm}
\subsection{Revolving Test}
\vspace{-1mm}
\noindent \textbf{Robustness to different azimuths.}
The detection robustness to different azimuths is important to the autonomous driving system, since sometimes the vehicle may turn at a large angle. For example, at small roundabouts or corner roads, the turning angle of the vehicle becomes large, resulting in a great change in the camera orientation. The autonomous vehicle should be able to accurately detect the surrounding objects even in such situations. To verify the robustness of the detector, we propose a revolving test to simulate this scenario: we turn the vehicle 60 degrees clockwise, namely, the original views/images [`FRONT LEFT', `FRONT', `FRONT RIGHT', `BACK RIGHT', `BACK', `BACK LEFT'] become the revolved ones [`BACK LEFT', `FRONT LEFT', `FRONT', `FRONT RIGHT', `BACK RIGHT', `BACK'], as illustrated in Figure~\ref{demo:revolving}. Then we evaluate the detector in the revolved view.

Compared with the original view, the performance of BEVDepth is degraded by 4.6\% NDS in the revolved view, as shown in Table~\ref{tab:revolving}. In particular, the most prominent degradation is in mAOE (by 15.6\%) and mAVE (by 22.7\%). In contrast, AeDet achieves a similar performance (47.3\% NDS \vs 47.1 \% NDS) in both views. This validates the robustness of AeDet with respect to the changes in azimuths, thanks to the azimuth-invariant feature learning and the azimuth-irrelevant target prediction. The predictions from the two views are displayed in Figure~\ref{demo:revolving}. AeDet yields almost the same prediction in both views.

\vspace{-1mm}
\subsection{Comparison with the State-of-the-Art}
\vspace{-1mm}
We compare our AeDet with other multi-view 3D object detectors on both the nuScenes \emph{val} set and \emph{test} set. The models are trained with CBGS strategy \cite{zhu2019class} for 20 epochs.
The performance on the nuScenes \emph{val} set is shown in Table~\ref{tab:val}.
For a fair comparison, we report the results at single-model single-scale testing. With ResNet-50 and ResNet-101, AeDet achieves 50.1\% NDS and 56.1\% NDS, outperforming the most current multi-view 3D object detectors such as BEVFormer \cite{li2022bevformer} (by 4.4\% NDS) and BEVDepth \cite{li2022bevdepth} (by 2.6\% NDS).
For the submitted results on the \emph{test} set, we use the \emph{train} set and \emph{val} set for training. We adopt ConvNeXt-B as the image backbone, and report the results of single model with test-time augmentation. As Table~\ref{tab:test} shows, AeDet sets a new state-of-the-art result with 53.1\% mAP and 62.0\% NDS in multi-view 3D object detection.

\vspace{-1mm}
\section{Conclusion}
\vspace{-1mm}
In this work, we propose an azimuth-equivariant detector AeDet, which is able to perform azimuth-invariant 3D object detection. Specifically, we first rotate the sampling grid of the convolution along the azimuth to extract the azimuth-invariant features. Then we rotate the anchor to allow the detection head to predict the azimuth-irrelevant targets. In addition, we further introduce a camera-decoupled virtual depth to unify the depth prediction for the images from different cameras.
With these improvements, AeDet sets a new state-of-the-art result with 62.0\% NDS on nuScenes.
\clearpage

{\small
\bibliographystyle{ieee_fullname}
\bibliography{egbib}
}

\clearpage
\begin{figure*}[t]
    \centering
    \includegraphics[width=\textwidth]{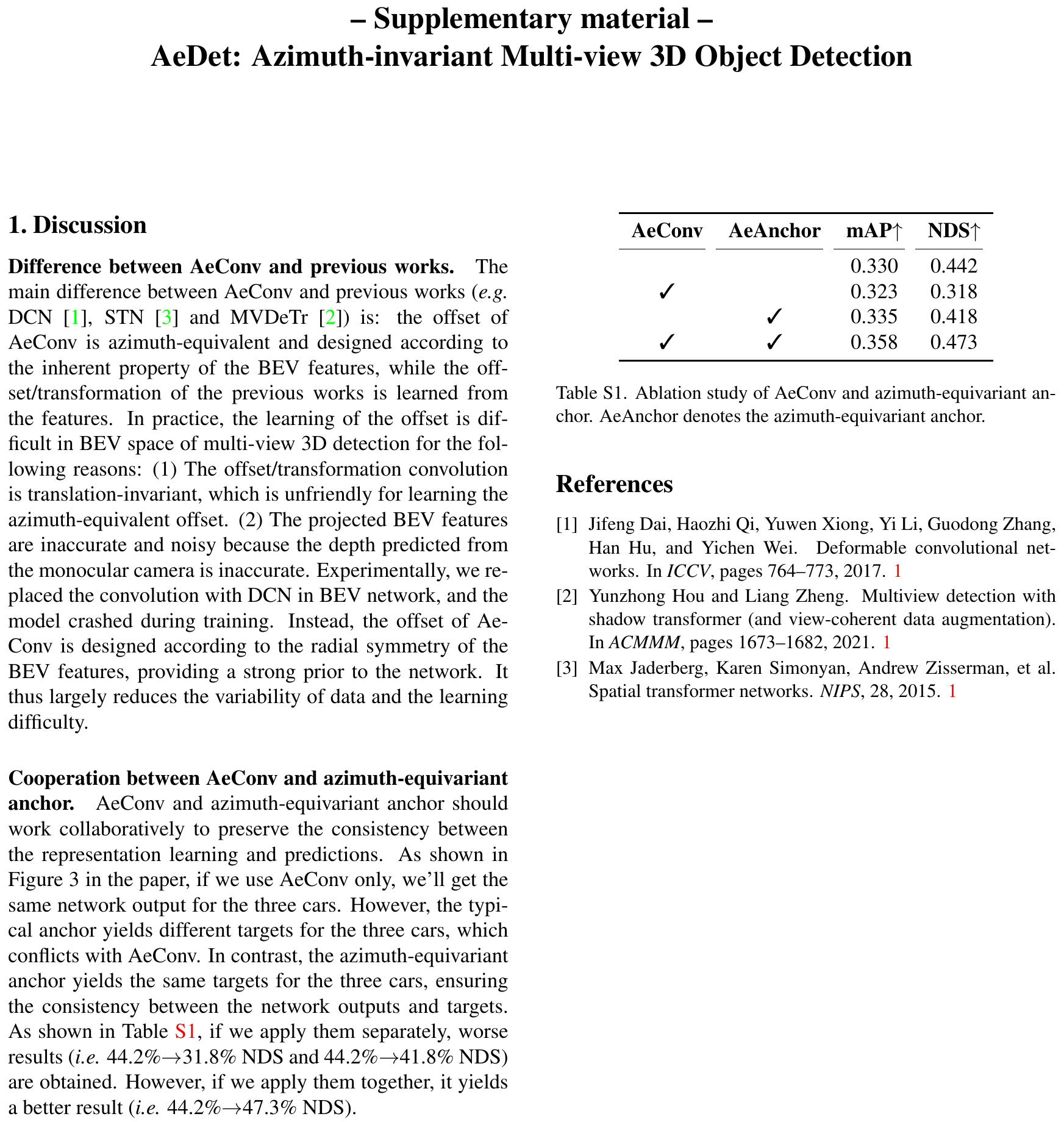}
\end{figure*}

\end{document}